\documentclass[encoding=utf8]{article}
\usepackage[a4paper, margin=1in]{geometry}

\usepackage{csquotes}
\usepackage[square,numbers]{natbib}
\usepackage{scrhack}
\usepackage{tabularx}
\usepackage{blindtext}
\usepackage{caption}
\usepackage{imakeidx}
\usepackage{siunitx}
\usepackage{graphicx}
\usepackage{epstopdf}
\usepackage{subfigure}
\usepackage{paralist}
\usepackage{epigraph}
\usepackage{algorithm}
\usepackage{algpseudocode}
\usepackage{amsmath}
\usepackage{amssymb}
\usepackage{bm}
\usepackage{adjustbox}
\usepackage{float}
\usepackage[font={small}]{caption}
\usepackage[parfill]{parskip}
\usepackage{etoolbox}
\usepackage{url}

\title{Simulating Liquids with Graph Networks}
\author{Jonathan Klimesch, Philipp Holl, Nils Thuerey}
\date{\today}

\begin{document}

\maketitle
	
\begin{abstract}
    Simulating complex dynamics like fluids with traditional simulators is computationally challenging.
    Deep learning models have been proposed as an efficient alternative, extending or replacing parts of traditional simulators.
    We investigate graph neural networks (GNNs) for learning fluid dynamics and find that their generalization capability is more limited than previous works would suggest.
    We also challenge the current practice of adding random noise to the network inputs in order to improve its generalization capability and simulation stability.
    We find that inserting the real data distribution, e.g. by unrolling multiple simulation steps, improves accuracy and that hiding all domain-specific features from the learning model improves generalization.
    Our results indicate that learning models, such as GNNs, fail to learn the exact underlying dynamics unless the training set is devoid of any other problem-specific correlations that could be used as shortcuts.
\end{abstract}

\section{Introduction}  \label{chap:introduction}
Simulating complex dynamics is invaluable to many fields in computer graphics and the natural sciences.
Numerical simulations of such dynamics are computationally challenging and are often inaccurate in approximating the underlying physical systems~\cite{2018_Rasp}.
Recent work suggests, that deep learning models can be an efficient and accurate alternative to traditional, hand-crafted simulation approaches~\cite{2020_sanchez, 2019_He, 2018_Thuerey}.
Graph networks in particular have become popular in collider physics~\cite{2020_shlomi}, astrophysics~\cite{2020_cranmer} or chemistry~\cite{2017_gilmer} and have been used to build various data-driven simulators~\cite{2016_battaglia, 2017_gilmer, 2018_battaglia, 2018_sanchez, 2019_li}.

Recently, Sanchez-Gonzales \emph{et al.} proposed the \emph{Graph Network-based Simulators} (GNS) framework, where they approximate fluid dynamics by learned message-passing on particle graphs \cite{2020_sanchez}.
We replicate their results on data from our custom fluid-Implicit-Particle (FLIP) \cite{1986_Brackbill, 2005_zhu} simulator, and extend the GNS framework with new training variants, investigating the impact of the random noise and the generalization of the trained model to different domains.

We use the differentiable physics framework $\Phi_\mathrm{Flow}$~\cite{2020_Holl} to implement our FLIP simulator in a differentiable manner, allowing us to roll out multiple simulation steps during training, similar to~\cite{2020_um}.
This enables us to formulate an alternative training scheme where we feed the predicted next simulation state back to the network as an input.
The network can thus learn under conditions matching inference time, which makes the noise term superfluous.


\section{Network architecture}  \label{chap:learning}
\begin{figure} [t]
    \centering
    \includegraphics[width=.9\textwidth]{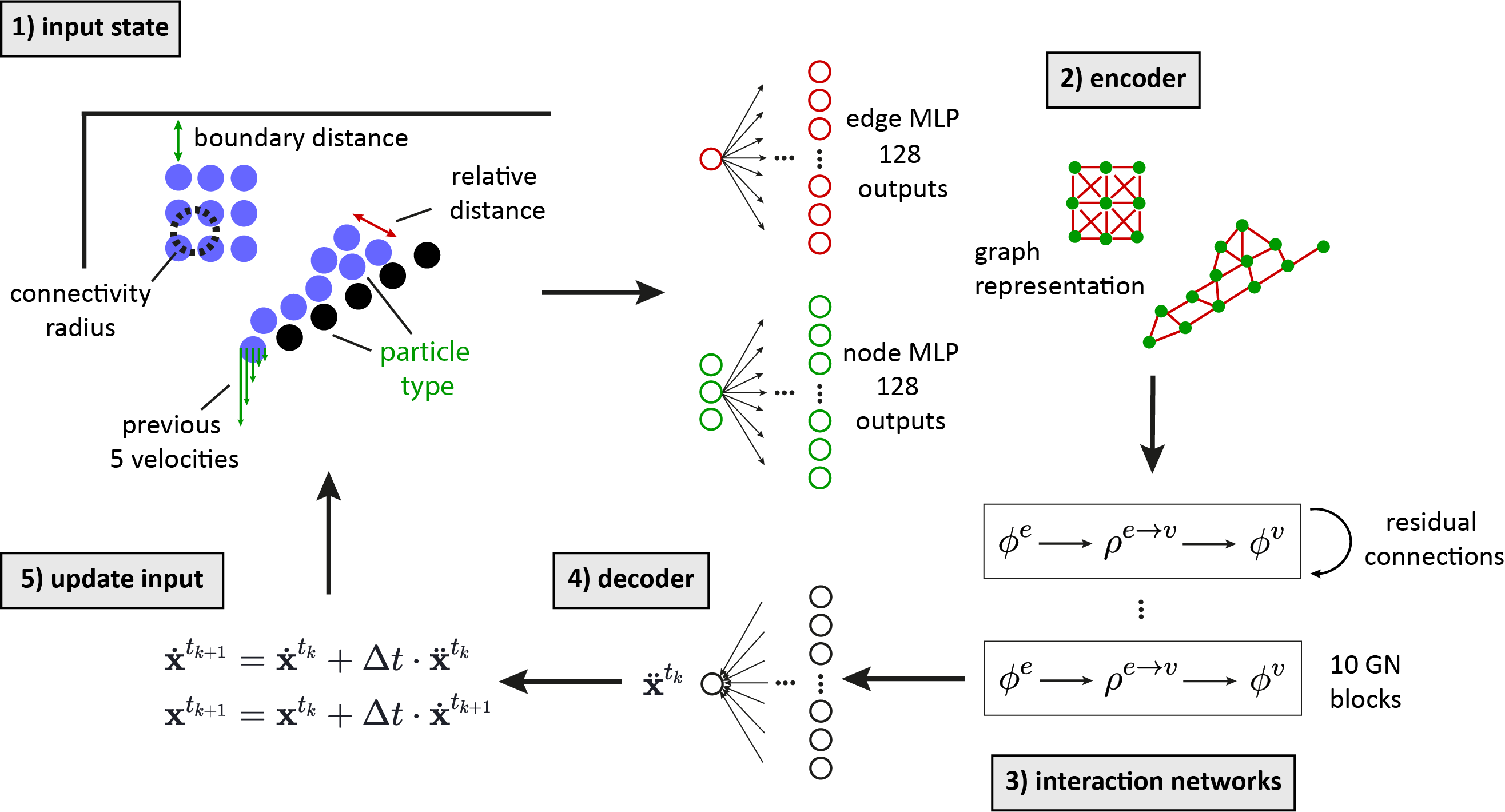}
    \caption{Schematic overview of the Graph Network Simulator (GNS) architecture.}
    \label{fig:gns}
\end{figure}

We adopt the Graph Network-based Simulators architecture from Sanchez-Gonzalez \emph{et al.} \cite{2020_sanchez} and train it on the dynamics of our FLIP simulator.
They compose several GN blocks into an encode-process-decode configuration which they name \emph{Graph Network-based Simulators} (GNS).
They demonstrate that the GNS is capable of learning complex dynamics of different materials, such as fluids or sand.
Figure \ref{fig:gns} displays the GNS architecture schematically.
On a high level, the GNS is a parameterization of dynamics which map the current state of the world to a future state. In case of the FLIP simulation, the dynamics are described by fluid dynamics.
In the Lagrangian view, the state of a fluid can be described in terms of single particles which interact with each other.
The GNS takes the positions of these particles from the FLIP simulation as input and uses its encoder to transform these positions into a graph, where each node corresponds to a particle, and edges indicate particles interacting with each other.
For each particle, edges are added to all neighboring particles within a certain \emph{connectivity radius}, which is a hyperparameter of the GNS architecture.
The node features are calculated by a multilayer perceptron (MLP) which takes the five previous particle velocities, the particle type and its distance to the domain boundaries as input, and outputs a latent vector of size 128.
The edge features are calculated by another MLP, which takes the relative distance between the connected nodes as input and produces another latent vector of size 128.

The particle interactions are then computed by performing message passing on the resulting graph.
This is done by a sequence of 10 different interaction networks, all having residual connections between the node and edge attributes at input and output.
The update functions of the interaction networks are implemented as MLPs and the aggregation function as an element-wise sum. The parameterization of the fluid dynamics thus takes the form of message-passing on a graph, performed by MLPs.

The resulting graph is then used to map the input state, the particle positions, to the next state. After the message passing, the GNS first uses its decoder, another MLP, to transform the node features into particle-wise accelerations. These accelerations are then used to calculate future velocities and positions for each particle using Euler integration (Figure~\ref{fig:gns}). 

All MLPs consist of two hidden layers with ReLU activations and size 128. The MLPs of the encoder use LayerNorm \cite{2016_Ba} after their output layer. A detailed analysis of the hyperparameters of the GNS architecture (e.g. number of velocities to calculate node features, number of interaction networks) can be found in the work of Sanchez-Gonzales \emph{et al.}~\cite{2020_sanchez}. They implemented the GNS architecture in TensorFlow 1 \cite{2016_abadi}, using Sonnet 1 \cite{2017_sonnet} and the Graph Nets library \cite{2018_graphnets}. The implementation was published on GitHub and is also used to apply different variants of the GNS to our FLIP simulation, as described in section \ref{sec:learning_variants}. 

\section{FLIP data sets} \label{sec:data set}
\begin{figure} [h]
    \centering
    \includegraphics[width=1\textwidth]{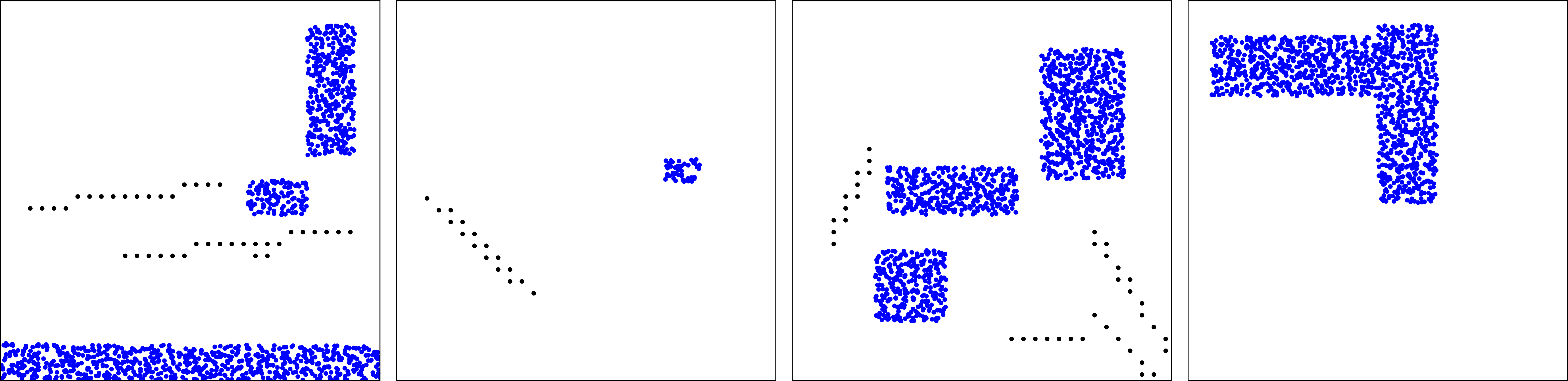}
    \caption{Simulation examples from our training set.}
    \label{fig:data set}
\end{figure}

\begin{figure} [h]
    \centering
    \includegraphics[width=1\textwidth]{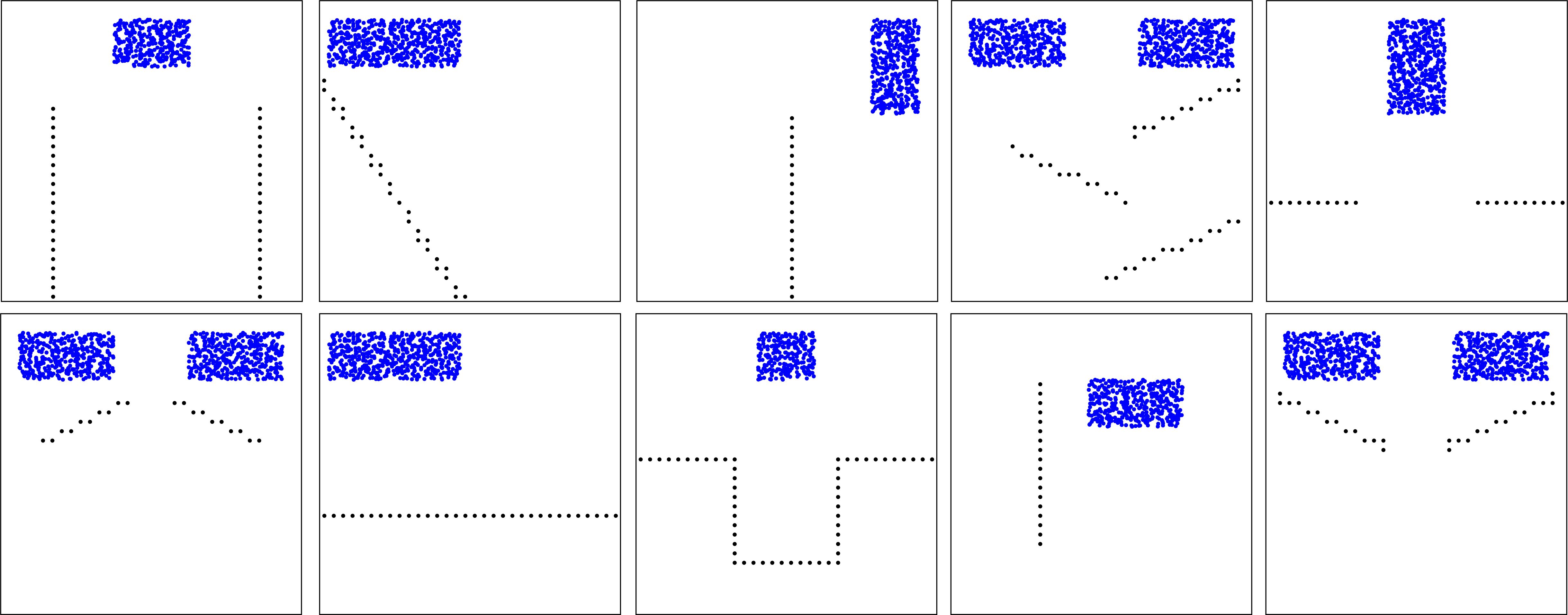}
    \caption{Simulation examples from our custom-designed test set.}
    \label{fig:test_set}
\end{figure}

To apply the GNS architecture to the FLIP simulation, we first use our simulator to generate training, validation and test sets.
Table \ref{tab:data set_specs} summarizes the parameters which we use to generate the training and validation sets. The training set consists of 2000 simulations with 400 frames each, using a time step of $0.05$ s. The scenes are randomly generated and can contain liquid blocks, elongated obstacles and a liquid pool at the bottom. Figure \ref{fig:data set} shows initial simulation states as examples of the training set. The validation set consists of 20 simulations with length 400 and is generated with the same random distribution as the training set, using a different random seed. We generate our test set manually to evaluate the behavior of the models in 10 challenging scenarios. Examples from the test set are shown in figure \ref{fig:test_set}.

\begin{table} [h]
    \centering
    \begin{tabular}{ |p{6.5cm}|p{1.5cm}| } 
    \hline
    \textbf{Parameter description} & \textbf{Value} \\
    \hline
    Trajectory duration (steps) & 400 \\
    \hline
    Time step (s) & 0.05 \\ 
    \hline
    Maximum number of particles & 1300 \\
    \hline
    Domain size & $32\times32$ \\ 
    \hline
    Probability for pool & 0.3 \\
    \hline
    Pool height & [3, 8] \\
    \hline
    Block size & [2, 20] \\
    \hline
    Probability for multiple blocks & 0.3 \\
    \hline
    Number of blocks if there are multiple & [2, 3] \\
    \hline
    Probability for obstacles & 0.8 \\
    \hline
    Number of obstacles & [1, 5] \\
    \hline
    Obstacle length & [2, 20] \\
    \hline
    Obstacle rotations & [0, 90] \\
    \hline
    Probability for initial velocity & 0.3 \\
    \hline
    Initial velocity range & [-5, 5] \\
    \hline
    \end{tabular}
    \caption{Parameter specifications for data set generation. Tuples indicate ranges from which values were drawn randomly. The use of some of the parameters is dependent on the corresponding probabilities (e.g. number of obstacles is only considered if probability for obstacles is met).}
    \label{tab:data set_specs}
\end{table}

Sanchez-Gonzales \emph{et al.} \cite{2020_sanchez} normalize all input and target vectors element-wise to zero mean and unit variance. Thus, we calculate the data set statistics online during data generation and use them later for normalization during training and inference. Due to memory constraints, we limit the number of particles in each FLIP simulation to less than $1300$. All trajectories are generated on domains of size $32 \times 32$ and scaled to size $0.8 \times 0.8$ with positions laying between $0.1$ and $0.9$.

The training set used by Sanchez-Gonzales \emph{et al.} \cite{2020_sanchez} is mostly generated by simulators based on the Material Point Method (MPM). They show that the connectivity radius is one of the most important hyperparameters of the GNS architecture. Their experiments indicate that greater connectivity radii yield lower errors. As explained in their work, larger particle neighborhoods enable long-range communication among the particles, supporting the message passing which applies the dynamics. However, this communication benefit stands in trade-off with the amount of computation and memory, which increases with larger radii due to the size of the graphs. 

\begin{figure} [h]
    \centering
    \includegraphics[width=1\textwidth]{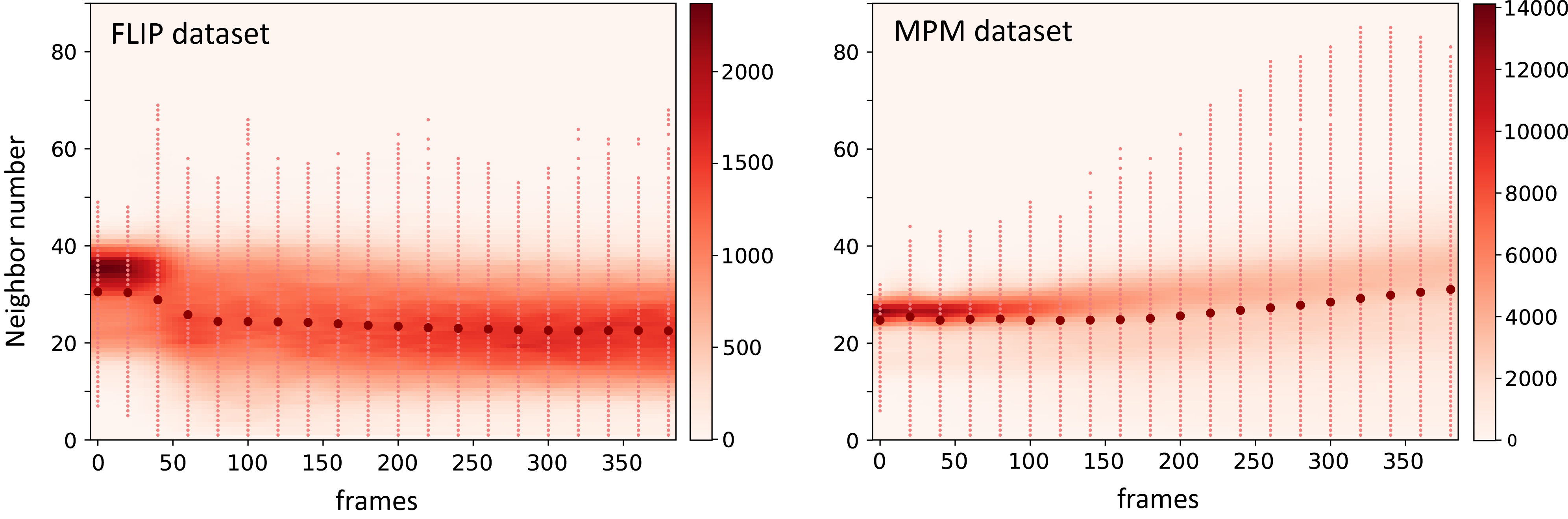}
    \caption{Distribution of neighbor numbers for the connectivity radius $0.03$ at different time steps, averaged over 50 data set samples. The left side shows results for the FLIP data set, the right side shows results for the MPM WaterRamps data set used in \cite{2020_sanchez}. The thick red dots mark the mean value of neighbor numbers. The distribution was interpolated between values at the positions of the thin red dots. The color indicates the number of particles with the respective number of neighbors.}
    \label{fig:radius}
\end{figure}

Sanchez-Gonzales \emph{et al.} use a connectivity radius of $0.015$ on a domain of size $0.8 \times 0.8$. Using this connectivity radius to train the GNS on FLIP trajectories yielded unstable results where the fluid blocks were torn apart within the first few time steps. Instead, a higher connectivity radius of $0.03$ was necessary in order to produce physical trajectories. Figure \ref{fig:radius} shows the time evolution of the distribution of neighboring particles (averaged over 50 simulations), using a connectivity radius of $0.03$ for the FLIP and MPM WaterRamps data set from \cite{2020_sanchez}. The left side of Figure \ref{fig:radius} shows that the neighbor number in our FLIP simulations drops at around frame number 50 indicating that the liquid blocks splash at the bottom. After that initial drop, the mean number of neighbor particles (thick red dots) stays nearly constant. This is expected due to the incompressibility constraints of the simulation. For the MPM trajectories (Figure \ref{fig:radius} right), the mean number of neighbor particles slightly increases, indicating that the fluid gets compressed over time. This enables long-range communications with a smaller radius. The maximum number of neighbor particles differs strongly between FLIP and MPM simulations, reaching higher values in the MPM case. This further indicates that even a low connectivity radius enables long-range communication between particles in the MPM case while a higher radius is required in case of our FLIP data set. 

\section{Training schemes and loss functions}
\label{sec:learning_variants}

Sanchez-Gonzales \emph{et al.} \cite{2020_sanchez} optimize the GNS parameters $\Theta$ using the $L_2$ loss 

\begin{equation}
    \mathcal{L} = \left\|\ddot{\mathbf{p}}_{\mathrm{GNS}}^t-\ddot{\mathbf{p}}_{\mathrm{GT}}^t\right\|^{2}
    \label{eq:one_step_loss}
\end{equation}

which compares the predicted accelerations (the outputs of the GNS) with the corresponding ground truth (GT) accelerations at time step $t$. A trained GNS can then be used to generate longer simulation rollouts by using the GNS outputs as its new inputs for the next time step. However, due to the complex mechanics and imperfect optimization, the GNS would then accumulate its own error over time. With the loss defined in equation \ref{eq:one_step_loss} the GNS model is only trained on one-step data and is therefore not trained for handling this accumulated error. One countermeasure used by Sanchez-Gonzales \emph{et al.} \cite{2020_sanchez} is to corrupt the model input during training with artificial noise, which ideally matches the noise produced by the model during rollout generation. 

First, we verify the capabilities of the GNS framework by training it on our FLIP data set. Except for the connectivity radius described in the last section, we train the first GNS (named 1-step-noise, 1sn) with the same procedure as used by Sanchez-Gonzales \emph{et al.} \cite{2020_sanchez}. Additionally, we train another GNS (named 1-step, 1s) without the additional artificial noise. 

As a next step, we implement an alternative to the artificial noise. This alternative uses a loss function which spans over more than one step and is described as

\begin{equation}
    \mathcal{L} = \frac{1}{n} \left( \left\|\ddot{\mathbf{p}}_{\mathrm{GNS}}^t(\mathbf{p}_{\mathrm{GT}}^{t-1})-\ddot{\mathbf{p}}_{\mathrm{GT}}^t\right\|^{2} + \sum_{i=1}^n \left\|\ddot{\mathbf{p}}_{\mathrm{GNS}}^{t+i}(\mathbf{p}_{\mathrm{GNS}}^{t+i-1})-\ddot{\mathbf{p}}_{\mathrm{GT}}^{t+i}\right\|^{2} \right)
    \label{eq:multi_step_loss}
\end{equation}

where $n$ is the number of additional steps using the GNS positions as input for the next prediction. The GNS first takes the five previous ground truth velocities as input ($\mathbf{p}_{\mathrm{GT}}^{t-1}$), outputs the particle-wise accelerations and calculates the new velocities and particle positions. In the next step, it takes the four previous ground truth velocities and the newly predicted velocities as input ($\mathbf{p}_{\mathrm{GNS}}^{t+i-1}$) and calculates the next velocities which replace another ground truth velocity in the next step. This way, the model is constantly confronted with its own error which should improve its error mitigation capabilities during rollout generation. Due to memory constraints, we test this alternative only with $n=1$. We optimize one GNS using this loss from scratch (named 2-step-scratch, 2ss) and another one by initializing it using pretrained weights from the 1-step model described above (named 2-step-initialized, 2si). 

As described above, the node features in the graph encoding are calculated by an MLP using the five previous velocities, the particle types and the distances to the domain boundaries as input. However, giving the GNS information about the domain boundaries restricts its generalization capabilities to the domain it has seen during training. An alternative which we explore in this report is to model the domain boundaries as obstacles and to remove the additional domain boundary distances from the node feature calculation. This alternative has two advantages. First, the model is not restricted to any specific domain, but operates solely on the interaction between fluid and obstacle particles. Second, the model sees more interactions between fluid and obstacle particles during training, which might improve its capabilities even further. We train a fifth model on a new version of our FLIP training set where we represent the domain boundaries as obstacle particles. We train this model with artificial noise, using the loss from equation \ref{eq:one_step_loss} and name it 1-step-noise-bounded (1snb). 

We optimize the parameters of our GNS models with the Adam optimizer \cite{2015_kingma} and a batch size of 2. The learning rate is decreased exponentially from $10^{-4}$ to $10^{-6}$. We adopted this optimization procedure from Sanchez-Gonzales \emph{et al.} and use it for all our experiments.

\section{Quantitative comparison}
\label{sec:comparison}

To compare the model variants from section \ref{sec:learning_variants} quantitatively, we use 4 different metrics. The first metric is the particle-wise mean-squared error of one-step acceleration predictions (MSE-acc 1) as defined in equation \ref{eq:one_step_loss}. The second metric (MSE 20) is the MSE averaged across time, particles and spatial dimensions of 20 frames, taken at each 20 steps of the full 400-step rollouts. This metric indicates the model performance at different stages of the FLIP trajectories (e.g. the falling liquid block in the beginning or the sloshing liquid at the bottom of the domain at later time steps). The third metric (MSE 400), is the MSE averaged across time, particle and spatial dimensions, but applied to the full rollouts of length 400.

Evaluating rollouts solely with particle-wise MSE can be misleading. Considering two particles A and B, the MSE could be high if the model predicts particle A at the position of particle B and vice versa, even so the particle distributions of prediction and ground truth match. Due to the chaotic nature of fluid motion, the MSE can therefore be misleading and one might favour a distributional metric, which is invariant under particle permutations. Therefore, we evaluate the model variants from section \ref{sec:learning_variants} with a fourth metric (EMD) using optimal transport (OT) and the earth mover's distance or Wasserstein metric \cite{2003_Villani, 2013_Cuturi}. 

The optimal transport problem describes an optimization problem of the form 

\begin{equation}
    d_{M}(\mathbf{r}, \mathbf{c})=\min _{P \in U(\mathbf{r}, \mathbf{c})} \sum_{i, j} P_{i j} M_{i j}.
    \label{eq:optimal_transport}
\end{equation}

Here, $\mathbf{r}$ and $\mathbf{c}$ describe the probabilistic weights (sum to 1) of all particles in the source and target distribution and $U(\mathbf{r}, \mathbf{c})$ is the set of positive $n \times m$ matrices defined by

\begin{equation*}
    U(\mathbf{r}, \mathbf{c})=\left\{P \in \mathbb{R}_{>0}^{n \times m} \mid P \mathbf{1}_{m}=\mathbf{r}, P^{\top} \mathbf{1}_{n}=\mathbf{c}\right\}.
\end{equation*}

$M$ is the cost matrix, which in case of a fluid with $n$ particles has the size $n \times n$ and contains the particle-wise distances. The optimization problem, described by equation \ref{eq:optimal_transport}, is to find the Wasserstein distance $d_M(\mathbf{r},\mathbf{c})$ which is calculated by using the element of $U(\mathbf{r}, \mathbf{c})$ which yields the lowest distance between the particle distribution of ground truth and prediction, measured by multiplying it with the cost matrix $M$. This distance is also called the \emph{earth mover distance} (EMD). This metric therefore relates each particle of the predicted particle distribution to its nearest neighbors in the target distribution and is invariant under particle permutations. For our evaluations, we calculate the Wasserstein distance using the Python Optimal Transport (POT) library \cite{2017_flamary}.

\begin{figure} [h]
    \centering
    \includegraphics[width=1\textwidth]{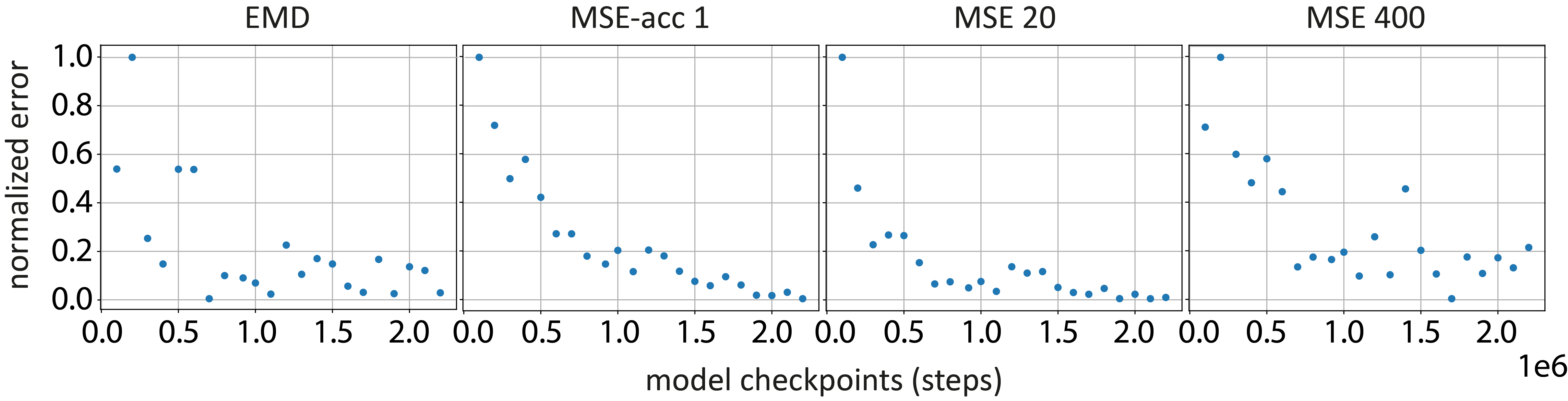}
    \caption{Screening results of the 1-step-noise model on the validation set during training.}
    \label{fig:model_screening}
\end{figure}

We evaluate the performance of each model on the validation set during training. Figure \ref{fig:model_screening} shows exemplary screening results using all four metrics for the 1-step-noise model variant described in section \ref{sec:learning_variants}. As the MSE-acc 1 is calculated in the same way as the loss with which this model was trained (equation \ref{eq:one_step_loss}) this metric shows a stable downwards trend with increasing number of training steps. The MSE 20 shows a similar behavior as it only includes 20-step rollouts and is still quite similar to the training loss. The MSE 400 and EMD metrics produce relatively noisy screenings which still show a strong downwards trend. The increase in noise is explained by the fact that the model is not directly optimized for the tasks measured by these metrics. Due to the chaotic weight updates within the network during training, small updates for the optimization with the one-step loss can have large impacts on the performance in the full trajectory rollout task.

\begin{table} [bt]
    \centering
    \begin{adjustbox}{width=\columnwidth,center}
    \begin{tabular}{ p{5cm}ccccc } 
    \hline
    Model variant & EMD ($10^{-2}$) & MSE-acc 1 ($10^{-1}$) & MSE 20 ($10^{-5}$) & MSE 400 ($10^{-2}$) \\
    \hline
    1-step & 2.598 & \textbf{3.775} & 5.653 & 2.853 \\
    1-step-noise & 1.617 & 4.980 & 5.574 & 2.469 \\
    1-step-noise-bounded & \textbf{1.336} & 4.985 & 5.648 & 2.959 \\
    2-step-scratch & 1.428 & 4.535 & 5.181 & 2.682 \\
    2-step-initialized & 1.367 & 3.933 & \textbf{4.711} & \textbf{2.430} \\
    \hline
    \end{tabular}
    \end{adjustbox}
    \caption{Testset performance of all model variants from section \ref{sec:learning_variants}, evaluated with multiple metrics. The MSE is reported for one-step (MSE-acc 1) acceleration predictions or averaged over either 20-step rollouts (MSE 20) or full rollouts (MSE 400). Bold values indicate the best score of the respective metric.}
    \label{tab:model_eval}
\end{table}

\begin{figure} [h]
    \centering
    \includegraphics[width=1\textwidth]{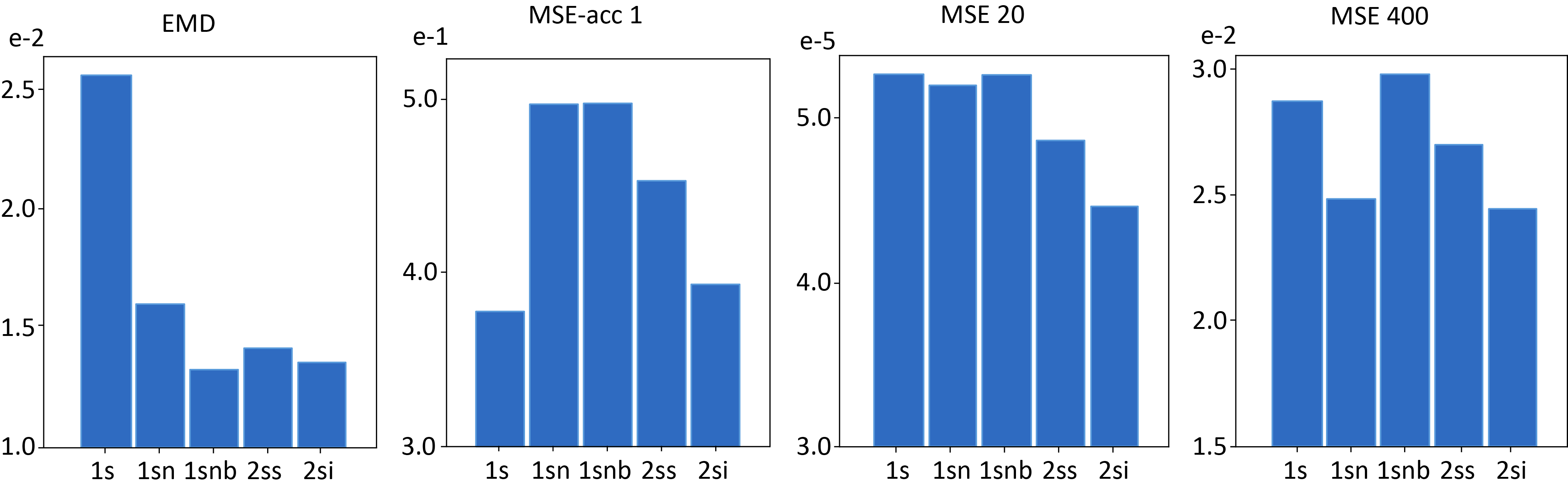}
    \caption{Test set performances of the 1-step (1s), 1-step-noise (1sn), 1-step-noise-bounded (1snb), 2-step-scratch (2ss) and 2-step-initialized (2si) models.}
    \label{fig:model_eval}
\end{figure}

\begin{figure}[H]
    \centering
    \includegraphics[width=.9\textwidth]{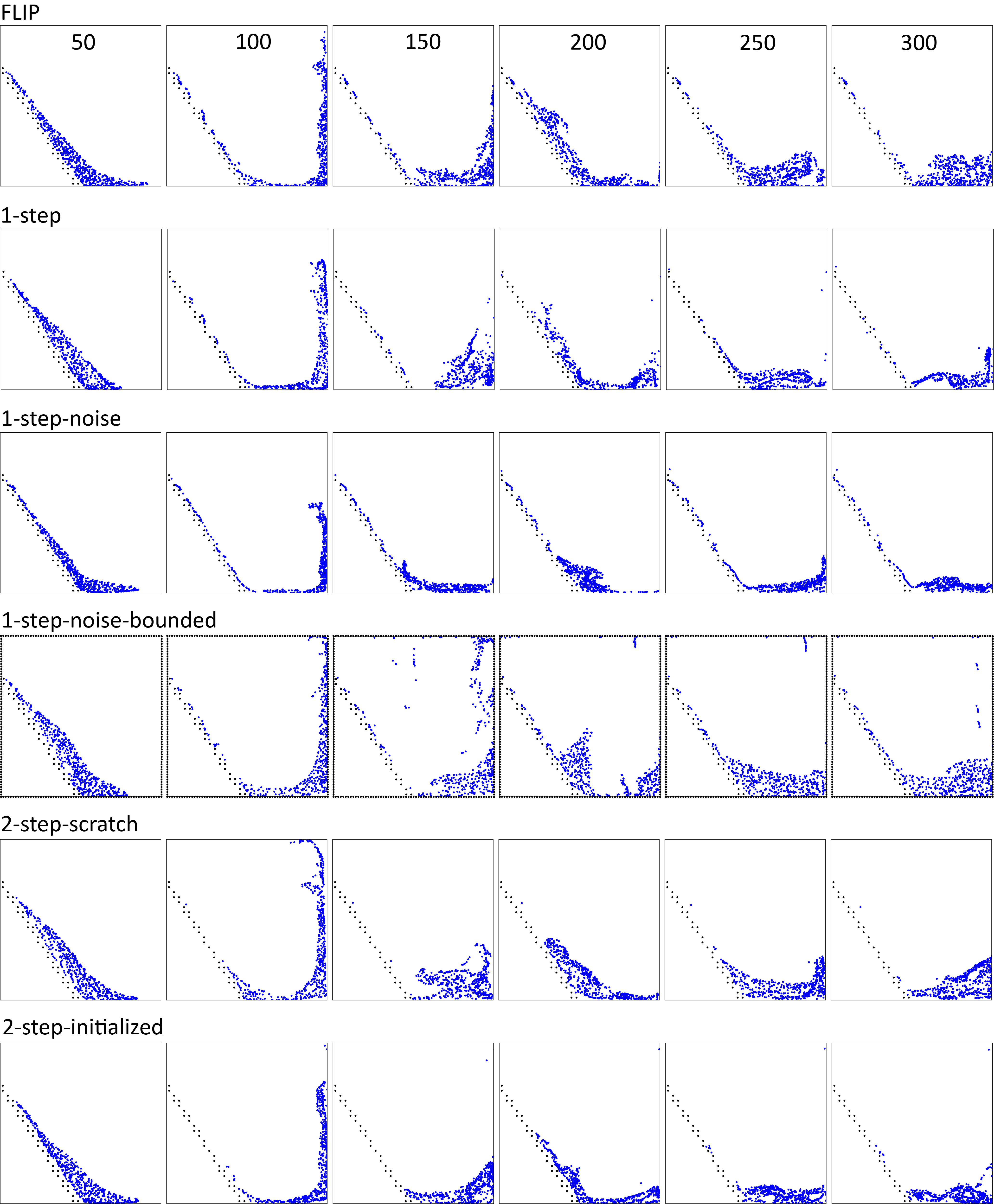}
    \caption{Example predictions for all five model variants. Different time steps are indicated at the top. The top row shows the ground truth trajectory. The initial state of the trajectory is shown at the top of the second column of figure \ref{fig:test_set}. Videos available at \protect\url{https://git.io/JOOQG}}.
    \label{fig:predictions}
\end{figure}

To favour models which have a high performance on generating full-scale rollouts, we use the MSE 400 to choose model checkpoints for further analysis and apply the best checkpoint of each model variant to the test set. Figure \ref{fig:model_eval} summarizes the performances of all models evaluated by the four metrics described before. The exact scores underlying this figure can be found in table \ref{tab:model_eval} in the appendix. Figure \ref{fig:trajectories} shows the trajectories of the MSE 400 and the EMD over time for all model variants, averaged over the entire test set. Figure \ref{fig:predictions} shows exemplary rollout predictions from all model variants for one of the test set scenarios.

\begin{figure} [h]
    \centering
    \includegraphics[width=1\textwidth]{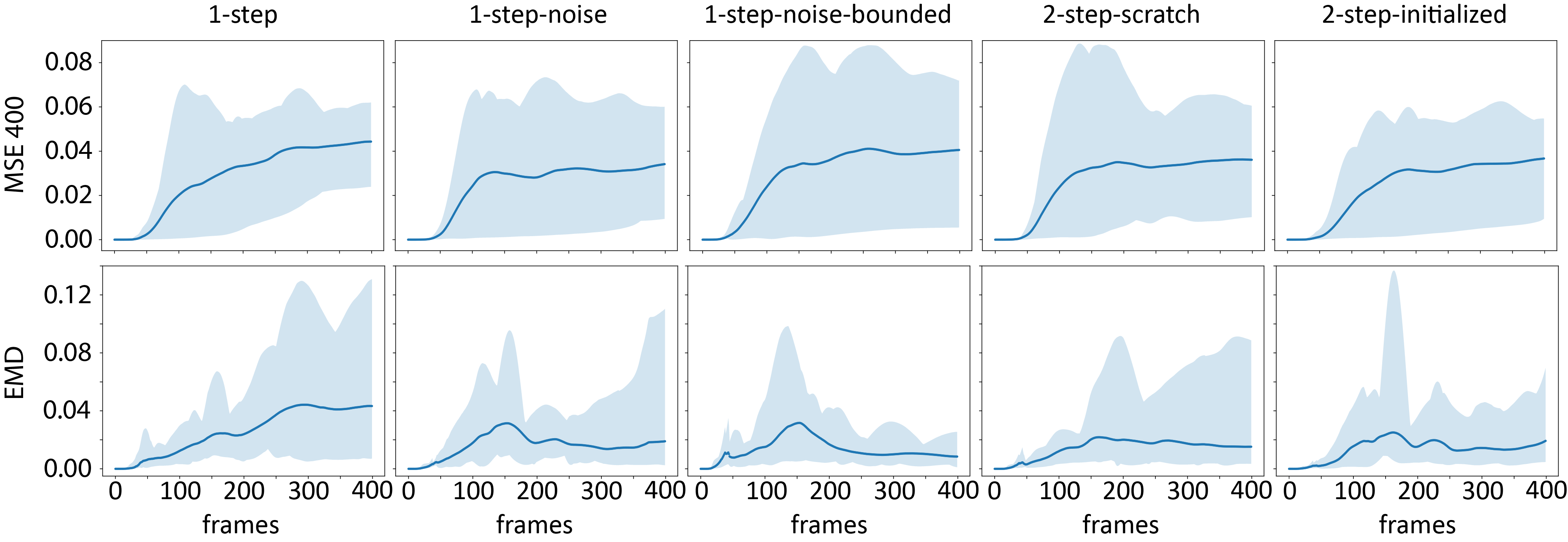}
    \caption{Error trajectories of the MSE 400 (MSE averaged over full rollouts) and the EMD metrics for all five model variants. The blue line represents the mean (over the entire data set) and the shaded area indicates the range of possible values.}
    \label{fig:trajectories}
\end{figure}

The 1-step model variant shows the best performance with respect to the MSE-acc 1 metric. This model is solely trained for this task, without any artificial noise or multi-step error, leading to its superior performance on the specific task of one-step predictions. While its MSE 20 and MSE 400 evaluations are rather similar to other model variants, its EMD evaluation shows a far worse performance. Therefore, the model's rollout prediction follows the ground truth rollout closely, but rather tries to hold the particles at similar positions, than to really apply dynamics which produce a similar visual behavior compared to the FLIP trajectory. The rollout example in figure \ref{fig:predictions} shows that the model sometimes produces strange, nonphysical motions (last column).

The 1-step-noise model shows a much better EMD performance, but performs worse than the 1-step model when evaluated with MSE-acc 1. This is likely due to the artificial noise which challenges this model with an additional difficulty during training. Figure \ref{fig:predictions} shows that this model variant leads to fluid motion which is more compressed than in the ground truth rollout. Similar compression can be seen in the rollout of the 1-step model. 

The 1-step-noise-bounded model has the best EMD performance, resembles the 1-step-noise model's performance for MSE-acc 1, but has lower performance in case of MSE 20 and MSE 400. With removal of the boundary distance features, the model has to extract this information solely from the interaction between fluid particles and obstacles. Figure \ref{fig:predictions} shows that this model variant produces rollouts with a similar density than the ground truth. However, it produces dynamics which cause the fluid to slide faster along the boundaries, possibly caused by the constant interactions with boundary particles which normally lets a fluid block splash at an obstacle.

For the 2-step model variants, the 2-step-initialized model has better performances in all metrics compared to the 2-step-scratch model. It surpasses all other variants on the MSE 20 and MSE 400 metrics and reaches similar performances to the best models for the MSE-acc 1 and EMD metrics. Thus, using a 1-step model as initialization for 2-step models seems to have a positive effect. However, figure \ref{fig:predictions} shows that it has inherited the compression tendency of the 1-step model, leading to a lower density of the fluid during rollout. In contrast, the 2-step-scratch model preserves the density throughout the rollout (similar to the 1-step-noise-bounded model). 

\begin{figure}[H]
    \centering
    \includegraphics[width=.8\textwidth]{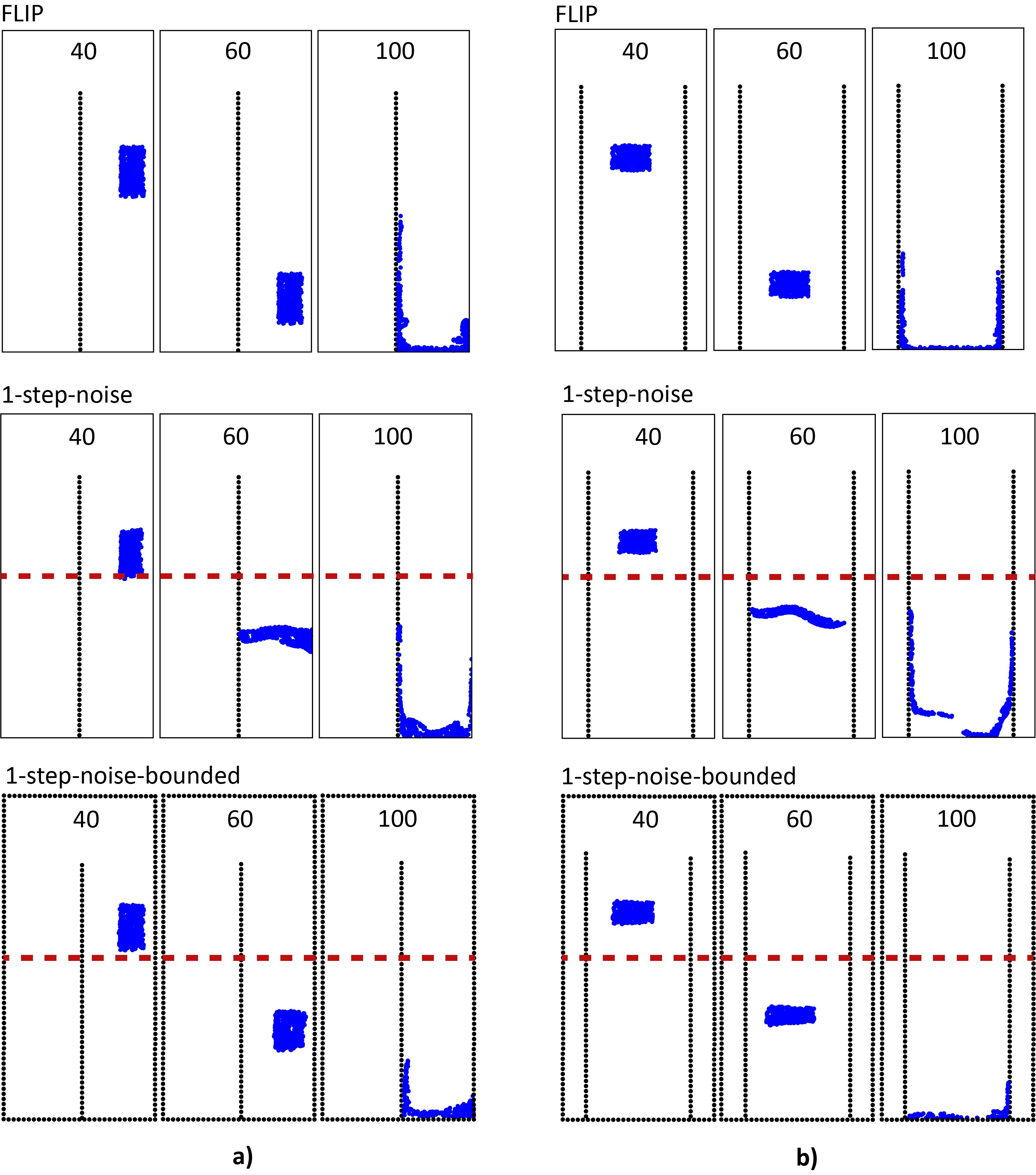}
    \caption{Two example predictions from the domain generalization experiment. Both left and right examples show ground truth (FLIP), predictions from 1-step-noise and predictions from 1-step-noise-bounded at the frames indicated by the numbers. The red dashed lines indicate the original domain size. Videos available at \protect\url{https://git.io/JOOQn}}.
    \label{fig:boundaries}
\end{figure}

As described in section \ref{sec:learning_variants} the 1-step-noise-bounded model should have the capability to generalize to larger domains. Figure \ref{fig:boundaries} shows two examples from such a generalization experiment. We extend the domain size of $32 \times 32$ to $32 \times 64$ and adapt the scaling to put the position values into the range of $0.1$ to $0.9$ in x and $0.1$ to $1.8$ in y direction. Then, we use both models, 1-step-noise and 1-step-noise-bounded (trained on $32 \times 32$ domains), to generate rollouts for these new scenarios. Figure \ref{fig:boundaries} shows that the 1-step-noise model produces dynamics which let the fluid splash when crossing half of the new domain. This height equals the bottom of the domain seen during training where fluid particles normally splash and are accelerated in x-direction. Therefore, the model seems to have learned to correlate accelerations in x-direction with the distance to the boundaries. 

The 1-step-noise-bounded model shows better behavior than the 1-step-noise model but still deforms the fluid block after crossing half of the domain. This indicates that the model normally not only correlates accelerations and boundary distances but also accelerations and velocities. In the training examples, most fluids had similar velocities right before splashing at the bottom. When the fluid crosses half of the new domain at height 32, the particles cross this velocity threshold, causing the model to decelerate the particles in y- and accelerate the particles in x-direction.

\section{Conclusions}  \label{chap:conclusions}
We have shown, that GNS models are capable of learning fluid dynamics from a dataset produced by our differentiable FLIP simulator.
Training the GNS models with our multi-step loss enables the models to mitigate accumulating errors in simulation rollouts and yields competitive results compared to models trained with the artificial noise proposed by Sanches-Gonzales \emph{et al.} \cite{2020_sanchez}.
Replacing the domain-specific boundary distance features with obstacle boundaries increases the generalization to larger domains.
However, we find that these models still have the tendency to deform the fluids as soon as they are passing the original domain size seen during training.
This indicates that the GNS does not learn the true underlying physics but rather problem-specific correlations between input velocities and output accelerations.
This is supported by the fact that the architecture takes the previous five velocities as inputs and does not only rely on just the previous velocity as one would expect from physical dynamics.
Furthermore, most models are unable to retain the original density of the fluid and compress the fluid particles much more than the FLIP simulator.

Extending the GNS architecture with strong inductive biases towards physical laws and symmetries could improve its physical understanding and force it to learn actual dynamics instead of problem-specific correlations.
Future work should also concentrate on examining the physical reasoning of learned simulators in more detail.
Transforming parts of learned simulators into symbolic models \cite{2020_cranmer, 2019_cranmer} and extending tools like the recently proposed GNNExplainer from Ying \emph{et al.} \cite{2019_Ying} could provide further insights into the reasoning of Graph Networks and could yield new ideas on how to improve their physical understanding.

\bibliographystyle{plain}
\bibliography{main}

\end{document}